\documentclass[10pt,twocolumn,letterpaper]{article}

\usepackage{wacv}
\makeatletter
\@namedef{ver@everyshi.sty}{}
\makeatother

\usepackage{times}
\usepackage{epsfig}
\usepackage{graphicx}
\usepackage{amsmath}
\usepackage{amssymb}
\usepackage{booktabs}

\usepackage{graphicx}
\usepackage{pifont}
\newcommand{\cmark}{\ding{51}}%
\newcommand{\xmark}{\ding{55}}%
\usepackage{tikz}
\usepackage{comment}
\usepackage[american]{babel}
\usepackage{amsmath,amssymb} 
\usepackage{color}
\usepackage{multirow}
\usepackage[accsupp]{axessibility}  
\usepackage{algorithm}
\usepackage{algorithmicx}
\usepackage{algpseudocode}

\newcommand{\squishlist}{
	\begin{list}{$\bullet$}
		{ \setlength{\itemsep}{0pt}
			\setlength{\parsep}{1pt}
			\setlength{\topsep}{1pt}
			\setlength{\partopsep}{0pt}
			\setlength{\leftmargin}{1.5em}
			\setlength{\labelwidth}{1em}
			\setlength{\labelsep}{0.5em} } }
\newcommand{\squishend}{\end{list} 
}
\def\wacvPaperID{} 

\wacvalgorithmstrack   

\wacvfinalcopy 

\ifwacvfinal
\usepackage[breaklinks=true,bookmarks=false]{hyperref}
\else
\usepackage[pagebackref=true,breaklinks=true,colorlinks,bookmarks=false]{hyperref}
\fi

\begin{document}

\title{A Simple and Efficient Pipeline to Build an End-to-End Spatial-Temporal Action Detector}

\author{Lin Sui$^2$\thanks{The work was done when Lin Sui was an intern at 4Paradigm.} \qquad Chen-Lin Zhang$^1$\thanks{Corresponding author} \qquad Lixin Gu$^3$ \qquad Feng Han$^3$ \\
$^1$ 4Paradigm Inc., Beijing, China \\
$^2$ State Key Laboratory for Novel Software Technology, Nanjing University, China \\
$^3$ DataElem Inc., Beijing, China \\
{\tt\small \{suilin0432, zclnjucs\}@gmail.com \; \{gulixin, hanfeng\}@dataelem.com}}

\maketitle
\thispagestyle{empty}

\begin{abstract}
Spatial-temporal action detection is a vital part of video understanding. Current spatial-temporal action detection methods mostly use an object detector to obtain person candidates and classify these person candidates into different action categories. So-called two-stage methods are heavy and hard to apply in real-world applications. Some existing methods build one-stage pipelines, But a large performance drop exists with the vanilla one-stage pipeline and extra classification modules are needed to achieve comparable performance. In this paper, we explore a simple and effective pipeline to build a strong one-stage spatial-temporal action detector. The pipeline is composed by two parts: one is a simple end-to-end spatial-temporal action detector. The proposed end-to-end detector has minor architecture changes to current proposal-based detectors and does not add extra action classification modules. The other part is a novel labeling strategy to utilize unlabeled frames in sparse annotated data. We named our model as SE-STAD. The proposed SE-STAD achieves around 2\% mAP boost and around 80\% FLOPs reduction. Our code will be released at \url{https://github.com/4paradigm-CV/SE-STAD}.
\end{abstract}

\section{Introduction}

Spatial-temporal action detection~(STAD), which aims to classify multiple persons' actions in videos, is a vital part of video understanding. The computer vision community has drawn much attention in the field of STAD~\cite{acrneccv2018,slowfasticcv2019,lfbcvpr2019}. 

In previous methods, STAD is often divided into two sub-tasks: actor localization and action classification. Previous methods mostly utilize a pre-trained object detector~\cite{fasterrcnnnips2015,resnextcvpr2017} and finetune it on the target dataset to obtain person candidates. Then, proposals are fed into the action classifier network to obtain the final action prediction. However, those two-stage methods are heavy and often need extra data~(such as MS-COCO~\cite{cocoeccv2014}). They need separate models and heavy computational resources. This prevents the current methods from real-world applications. A recent work~\cite{wooiccv2021} has shown that there is a dilemma between actor localization and action classification. Actor localization only needs a single image while action detection needs the whole input sequence. Thus, ~\cite{wooiccv2021} proposes an end-to-end method WOO, which uses a unified backbone to perform actor localization and action detection. However, they still 
have a significant performance drop with the vanilla structure and need to introduce an extra attention module into the classification head to enhance the performance.

\begin{figure}[t]
    \centering
    \includegraphics[width=0.45\textwidth]{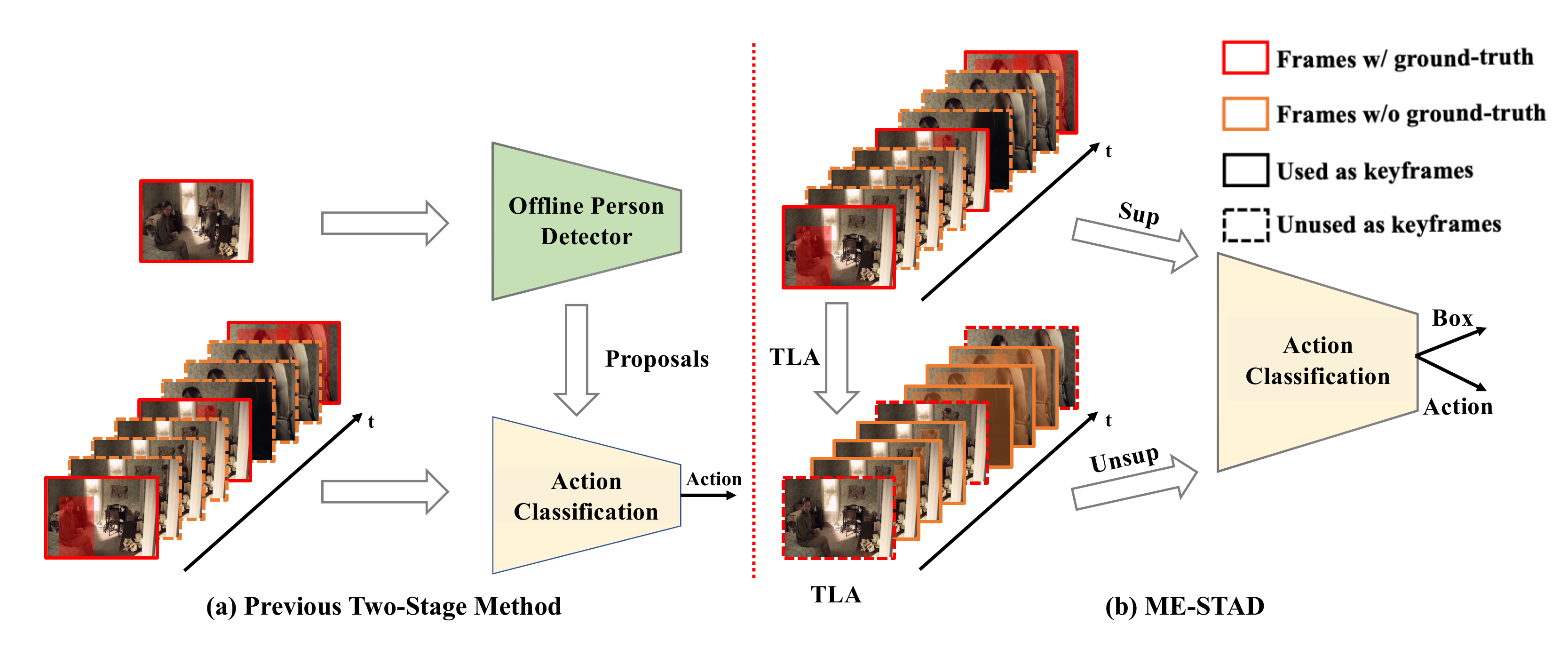}
    \caption{\textbf{Comparison between previous two-stage methods and our SE-STAD.} (a). Previous two-stage STAD methods use a heavy offline person detector, which also relies on additional data, to perform actor localization and they only use annotations of keyframes to train the action classifier. (b). Our SE-STAD trains an end-to-end spatial-temporal action detector in which actor localization part only occupies a small part of the computation. We also propose temporal label assignment~(TLA) to utilize unlabeled frames in large-scale sparsely annotated datasets like AVA~\cite{avacvpr2018}.}
    \label{fig:teaser}
\end{figure}

In this paper, we propose a new method named \texttt{Simple and Effective Spatial-Temporal Action Detection}, in short for SE-STAD. The general pipeline of SE-STAD is in Fig.~\ref{fig:teaser}. SE-STAD is composed by two parts. One is a strong one-stage spatial-temporal action detector architecture. Focusing on the localization ability of detector, the proposed architecture has minor architect modifications to existing two-stage methods. Thus, our methods can be applied to many existing STAD methods with comparable performance and less computational burden. With minor added components and effective training strategies, we empower the ability to conduct actor localization and action classification simultaneously without losing accuracy. Compared to existing works, our work is light-weighted and jointly optimized, which avoids the separate learning dilemma. Besides, we are the first to explore the strategy about building an end-to-end STAD from the perspective of localization ability. SE-STAD can also get benefits from other methods, such as adopting attention-based classification heads~\cite{wooiccv2021, acarcvpr2021}.

The second part is a new paradigm to utilize every possible information in sparsely annotated spatial-temporal datasets. Sparse annotation is an efficient and effective way to build a large-scale STAD dataset, only the keyframes will be annotated~(e.g. 1fps in AVA~\cite{avacvpr2018}). A huge amount of frames do not have annotations. Hence, we propose to utilize these unlabeled data to provide more clear temporal action boundaries and help the detector learn fine-grained information. Considering the distinctiveness of unlabeled data in sparse-annotated STAD datasets, we propose a novelty pseudo labeling strategy: temporal label assignment~(TLA) to generate pseudo labels. With the help of TLA, end-to-end spatial-temporal action detectors successfully enjoy performance gains from the neglected unlabeled data of sparse-annotated datasets.

Our contributions are listed as follows:

\squishlist
    \item We propose a simple and effective pipeline to build end-to-end spatial-temporal action detection methods. The proposed pipeline can be applied to many existing spatial-temporal action detection methods.
    \item We build a simple architecture for end-to-end action detection with effective training methods, which avoids extra offline person detector and achieves comparable performance with two-stage methods.
    \item A novel semi-supervised learning strategy with a pseudo temporal labeling strategy is proposed to utilize every possible information in sparsely annotated data. With the proposed pipeline, we achieve a 2.2\%/1.5\% mAP boost and around 80\%/20\% FLOPs reduction compared to both proposal-based methods and one stage methods with extra proposals. 
\squishend
\vspace{-0.3cm}
\section{Related Works}
\vspace{-0.2cm}

In this section, we will introduce works related to our SE-STAD, including spatial-temporal action detection, object detection and semi-supervised learning.

\textbf{Spatial-Temporal Action Detection} Spatial-temporal action detection~(STAD) aims to detect the action of different persons in the input video clips. Thus, STAD models needs to be aware of spatial and temporal information. After large-scale datasets are annotated and introduced~\cite{avacvpr2018,avakineticsarxiv2020}, researchers have paid much attention to STAD. 

Most existing works often follow a traditional Fast-RCNN~\cite{fastrcnniccv2015} pipeline with pre-extracted proposals to perform STAD~\cite{avacvpr2018,rtpeccv2018,lfbcvpr2019,acarcvpr2021,mviticcv2021}. A previous work~\cite{contextrcnneccv2020} shows the original R-CNN~\cite{rcnncvpr2014} pipeline works better for spatial-temporal action detection. However, those works are heavy and inefficient. 

Besides those two-stage methods, researchers also proposed some single-stage methods for action detection. 
Some works~\cite{avacvpr2018,avabaselinearxiv2018} also employ the Faster-RCNN~\cite{fasterrcnnnips2015} pipeline but with low performance. Early works including YOWO~\cite{yowoarxiv2019}, ACRN~\cite{acrneccv2018} and Point3D~\cite{point3dbmvc2021} combine pre-trained 2D and 3D backbones to build pseudo end-to-end detectors. Recently, WOO~\cite{wooiccv2021} proposes a single-stage, unified network for end-to-end action detection. WOO first utilizes Sparse-RCNN~\cite{sparsercnncvpr2021} along with the key frame to generate action candidates. Then action candidates will be fed into the classifier to obtain final results. With the vanilla structure, WOO has a major performance gap with the proposal-based methods. Thus, WOO utilizes an extra attention module to boost performance. In contrast, SE-STAD has a simple modification to the current proposal-based architecture. We only add a simple object detector and utilize better training strategies, and we achieve better results than WOO and proposal-based methods.

Apart from the detection structure side, Many researchers also propose new modules to enhance the performance including feature banks modules~\cite{lfbcvpr2019,acarcvpr2021}, attention modules~\cite{acarcvpr2021, wooiccv2021} and graph-based methods~\cite{vatcvpr2019,acrneccv2018,structcvpr2019}. However, we want to build a simple and strong model for end-to-end spatial-temporal action detection. Thus, we do not add any extra modules to our SE-STAD.

\textbf{Object Detection} Actor localization needs to detect the location of persons in the input image. Thus, object detection is needed. Object detection has been a popular area in the computer vision community. Early works often use a two-stage pipeline with pre-defined anchors~\cite{rcnncvpr2014,fastrcnniccv2015,fasterrcnnnips2015}. One-stage methods, especially anchor-free detectors are proposed to reduce the computational burden for object detection~\cite{retinaneticcv2017,yolocvpr2016,ssdeccv2016, fcosiccv2019, centernetarxiv2019}. Anchor-free detectors are easy to use in real-world applications. More recent methods want to train object detectors in an end-to-end manner~\cite{detreccv2020,sparsercnncvpr2021}.

In this paper, we adopt the one-stage anchor-free detector FCOS~\cite{fcosiccv2019}, as the person detector in SE-STAD. FCOS is a simple and effective choice for the person detector. 

\textbf{Semi-Supervised Learning}
Semi-supervised learning~(SSL) aims to achieve better performance with the help of additional unlabeled data. In brief, recent semi-supervised learning follows two main ways: introducing consistency regularization~\cite{pimodelnips2015,mixmatchnips2019,udanips2020} or performing pseudo-labeling~\cite{pseudolabelicmlworkshop2013,meanteachernips2017,r2d2aaai2020}. Some other works, such as~\cite{fixmatchnips2020} also combine these two ways into a single method. 

Semi-supervised object detection~(SSOD), which has received lots of attention recently, is an important subdomain in SSL. CSD~\cite{csdnips2019} used the consistency of predictions and proposed background elimination. Some other works~\cite{unbiasediclr2021,ismtcvpr2021,softteachericcv2021} built variants of the Mean Teacher~\cite{meanteachernips2017} framework and achieved promising performance gains.

However, in traditional semi-supervised learning tasks, the unlabeled data is introduced additionally without restrictions. Whereas, in large-scale sparsely annotated STAD datasets such as AVA~\cite{avacvpr2018}, the unlabeled frames have high correlations with the nearby labeled frames. Temporal restriction between the labeled part and the unlabeled part has not been explored in the STAD field yet.

\section{Methods}
In this section, we will give a detailed description of our SE-STAD.
\subsection{Notations}
We will first define the notations used in this paper.

Given a spatial-temporal action detection dataset $\mathcal{D}$, which is composed of a total number of $m$ videos: $\mathcal{D} = \{V_1,\ldots, V_m\}$. For simplicity, we suppose all videos in $\mathcal{D}$ have same height $h$, width $w$ and number of frames $n$. Thus, $V_i \in \mathbb{R}^{n \times h \times w \times 3}$. Spatial temporal action detection needs to detect the action category of the person in the specific input frame. For a frame $F_j$ in $V_i$, we need to detect a tuple $(x_1, y_1, x_2, y_2, \mathbf{cls})$ for each person in $F_j$. $(x_1, y_1, x_2, y_2)$ is the spatial location of the person, and $\mathbf{cls} \in [0, 1]^C$ is the action category where $C$ is the pre-defined set of action classes.
In widely used sparse-annotated datasets such as AVA~\cite{avacvpr2018}, ground-truth annotations are annotated at one frame per second. For such a dataset $\mathcal{D}$, We denotes the labeled part as $\mathcal{D}^l = \{V_1^l, \cdots, V_m^l\}$ with annotations $Y^l=\{A_1^l, \cdots, A_m^l\}$ and the unlabeled part as $\mathcal{D}^u = \{V_1^u, \cdots, V_m^u\}$.

\subsection{Motivation}
As shown in Sec.~2, previous STAD methods often follow a two-stage pipeline and utilize two networks: First conducting person detection with an offline object detector, then detected area of interests~(RoIs) will be fed into a traditional Fast-RCNN style network to obtain the final action predictions. The two-stage networks are not efficient. Besides, they always need extra data~(such as MS-COCO~\cite{cocoeccv2014}) to train the additional person detector. WOO~\cite{wooiccv2021} uses a unified backbone to perform actor localization and action classification simultaneously. However, their unified models result in a large performance drop and WOO~\cite{wooiccv2021} proposes an extra embedding interaction head to boost the performance. 

In contrast to those extra modules, we want to build a unified, end-to-end and simple method for spatial-temporal action detection. Thus, we want to make minimal modifications to current proposal-based methods, to perform spatial-temporal action detection effectively and efficiently. We name our proposed model as \texttt{Simple and Effective Spatial Temporal Action Detection~(SE-STAD)}.

\subsection{SE-STAD}

\begin{figure*}[t]
    \centering
    \includegraphics[width=0.75\textwidth]{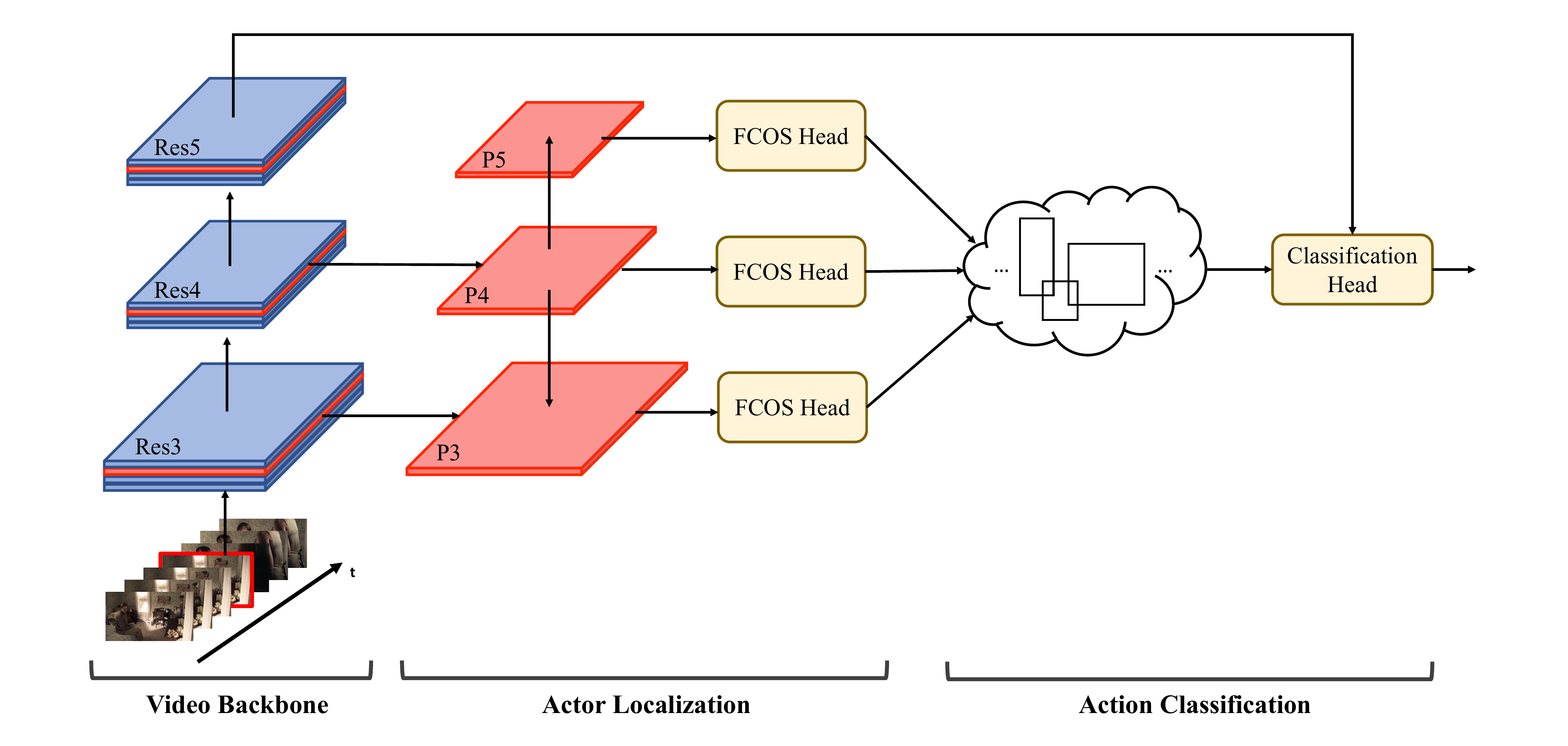}
    \caption{\textbf{Overview of our SE-STAD.} The whole pipeline consists of three parts: video backbone, actor localization part and action classification part. In the actor localization part, we build the feature pyramid on top of keyframes feature from Res3 and Res4 layers. After performing actor localization, proposals generated by FCOS heads will be used to extract features from Res5 and perform action classification.}
    \label{fig:architecture}
    \vspace{-1.5em}
\end{figure*}

Our proposed SE-STAD is composed of three parts: feature extraction part, actor localization part and action classification part. We unified the three parts into a single network. We will introduce these three parts step by step.

\subsubsection{Feature Extraction} 
In this part, we directly use existing action classification backbones, i.e., SlowFast~\cite{slowfasticcv2019} for feature extraction. Moreover, SE-STAD can utilize any modern backbones to boost performance, including the recent Transformer-based models, i.e., Video-Swin~\cite{videoswimarxiv2021}, ViViT~\cite{viviticcv2021} and MViT~\cite{mviticcv2021}.

\subsubsection{Actor Localization Part} 
We need to perform actor localization in our SE-STAD. Previously, separate pre-trained object detectors are adopted on actor localization, the most commonly used are Faster-RCNN~\cite{fasterrcnnnips2015} with a ResNeXt-101~\cite{resnextcvpr2017} backbone. However, the separate object detector has an extra heavy computational burden, which is inefficient. A recent work WOO~\cite{wooiccv2021} proposes to integrate an existing object detection head, i.e., Sparse R-CNN~\cite{sparsercnncvpr2021} into the current action classification backbone. In this paper, we follow the suggestions in WOO~\cite{wooiccv2021}, performing actor localization with the spatial feature of keyframes. Regarding the low input resolution and efficiency issue, we choose a popular one-stage anchor-free object detector FCOS~\cite{fcosiccv2019}. Thus, the loss for actor localization is:
\begin{equation}
    \mathcal{L}_{al} = \mathcal{L}_{cls}(\mathbf{c}_i, \mathbf{b}_i) + \mathcal{L}_{iou}(\mathbf{c}_i, \mathbf{b}_i) + \mathcal{L}_{centerness}(\mathbf{c}_i, \mathbf{b}_i)
\end{equation}
where $\mathbf{c}_i$ indicates the video clip, $\mathbf{b}_i$ means ground-truth bounding boxes of the keyframe of $\mathbf{c}_i$. $\mathcal{L}_{cls}$, $\mathcal{L}_{iou}$ and $\mathcal{L}_{centerness}$ are the Focal loss~\cite{retinaneticcv2017} for binary classification (existing of actor), GIoU~\cite{gioulosscvpr2019} loss for bounding box regression and centerness prediction, respectively. Compared to Sparse-RCNN~\cite{sparsercnncvpr2021}, FCOS has dense output proposals (before post-processing), and later we show that dense outputs~(without post-processing) matters in STAD performance. In Sec.~4, We ablate different heads for actor localization including anchor-based heads and anchor-free heads to verify the effectiveness of different heads, and the training strategy. Our models use a simple actor localization head and perform better than vanilla WOO~\cite{wooiccv2021}, even comparable to or better than WOO with extra attention modules. 

\subsubsection{Action Classification Part}

For action classification, since we want to build an end-to-end spatial-temporal action detection network with minimum efforts, we follow the common practice to build an action classification head: we use the traditional ROIAlign~\cite{retinaneticcv2017} layer with temporal pooling to get the feature for each actor proposal, then a simple linear layer is attached to get the final action predictions, we use the binary cross entropy loss to train the action classification head. Thus, the loss of the action classification head becomes:

\begin{equation}
    \mathcal{L}_{ac} = \mathcal{L}_{bce}(\mathbf{c}_i, \mathbf{b}_i, \mathbf{l}_i)
\end{equation}
where $\mathbf{l}_i$ denotes the classification annotation and $\mathcal{L}_{bce}$ is binary cross entropy loss. In order to balance the scale of localization loss $\mathcal{L}_{al}$ and classification loss $\mathcal{L}_{ac}$, we introduce loss weight $\lambda_{cls}$ for action classification which is set to 10 as default. Experiments in Sec.~4 show the model is robust to different $\lambda_{cls}$.

The overall structure of our SE-STAD is quite simple. We only add a simple FCOS head to perform actor localization. However, a simple model achieves comparable results to proposal-based methods~\cite{slowfasticcv2019} and the recently unified backbone method~\cite{wooiccv2021} which introduces additional attention modules. 

Besides the model structure, we propose a novel semi-supervised training strategy to better utilize every possible piece of information in the training video. With the semi-supervised training stage, our model can achieve better results than originally trained models. 
\subsection{Semi-Supervised Action Detection for SE-STAD}
It's well known that sparse annotation is an efficient and effective way to build large-scale spatial-temporal action detection datasets. However, as large parts of data are unlabeled, sparse annotations fail to provide clear temporal action boundaries. This phenomenon has been shown by previous literature~\cite{multisportsiccv2021}. Utilizing the unlabeled part is a natural way to help the detector to learn fine-grained information. Hence, we propose a new semi-supervised training method for sparsely annotated datasets in spatial-temporal action detection. 

When performing semi-supervised training in SE-STAD, we adopt the online updating paradigm. Besides, in order to avoid the inductive bias introduced in semi-supervised training, following the widely used Mean Teacher~\cite{meanteachernips2017} pipeline, we also build a teacher-student mutual-learning paradigm. Firstly, to get a good initialization for the detector, we do not perform semi-supervised training directly at first. That means we only use data with annotations to warm up the detector $D$ by Eq.~\ref{eq:burnin}.
\begin{equation}
    \mathcal{L}_{sup} = \frac{1}{N} \sum_{i} \mathcal{L}_{ac}(\mathbf{c}^s_i, \mathbf{b}^s_i, \mathbf{l}^s_i) + \lambda_{cls} \mathcal{L}_{al}(\mathbf{c}^s_i, \mathbf{b}^s_i)
    \label{eq:burnin}
\end{equation}

After warming up the spatial-temporal action detector $D$, weights of $D$ will be copied to the teacher model $D_{teacher}$ and student model $D_{student}$ as the initialization weights. Then we use both the labeled data and unlabeled data to further train the detector. The student model is updated via gradient back-propagation, but the gradient back-propagating to the teacher model is stopped. The teacher model is maintained by exponential moving average so as to eliminate the influence of inductive bias and provide more accurate person proposals for the student at the beginning. Loss function in this stage consists of losses on labeled data $\mathcal{L}_{sup}$~(Eq.~\ref{eq:burnin}) and unlabeled data $\mathcal{L}_{unsup}$~(Eq.~\ref{eq:unsup}).
\begin{equation}
    \mathcal{L} = \mathcal{L}_{sup} + \lambda_{unsup} \mathcal{L}_{unsup}
\end{equation}
\begin{equation}
    \mathcal{L}_{unsup} = \frac{1}{N} \sum_{i} \mathcal{L}_{ac}(\mathbf{c}_i^u, \mathbf{b}_i^u, \mathbf{l}_i^u) + \lambda_{cls} \mathcal{L}_{al}(\mathbf{c}_i^u, \mathbf{b}_i^u)
    \label{eq:unsup}
\end{equation}
where $\mathbf{b}_i^u$ and $\mathbf{l}_i^u$ are pseudo ground-truth annotations dynamically generated by temporal label assignment~(TLA) which will be discussed later.

\begin{algorithm}[t!]
    \caption{Temporal Label Assignment~(TLA)}
    \begin{algorithmic}[1]
        \Statex {\textbf{Input}: video clip $\mathbf{c}^u$, boxes and labels of nearest former keyframe $\mathbf{b}^{left}, \mathbf{l}^{left}$ and nearest later keyframe $\mathbf{b}^{right}, \mathbf{l}^{right}$}, detector $D$
        \Statex {\textbf{Output}: pseudo bounding boxes $\mathbf{b}^{u}$, pseudo lables $\mathbf{l}^{u}$}
        \State \quad {$\mathbf{b}^{u}, \mathbf{s} = D(\mathbf{c}^u)$}
        \State \quad {$\mathbf{b}^{gt} = \mathbf{b}^{left} \bigcup \mathbf{b}^{right}$, $\mathbf{l}^{gt} = \mathbf{l}^{left} \bigcup \mathbf{l}^{right}$}
        \State \quad \textbf{for} {$i=1, \ \ldots, \ N$}
        \State \quad \quad \textbf{for} {$j=1, \ \ldots, \ M$}
        \State \quad \quad \quad {$Cost_{ij} = \mathcal{L}_{bce}(\mathbf{s}_i, \mathbf{l}^{gt}_j) + \mathcal{L}_{L1}(\mathbf{b}_i^{u}, \mathbf{b}^{gt}_j)$} 
        \Statex \quad \quad \quad \quad \quad \quad {$ + \ \mathcal{L}_{iou}(\mathbf{b}_i^{u}, \mathbf{b}^{gt}_j)$}
        \State \quad {Assignment $\hat{\pi} = \arg\min_{\pi \in \prod_N^M} \sum_{i}^{N} Cost_{i, \pi(i)}$}
        \State \quad {$inds = [\hat{\pi}(1), \cdots, \hat{\pi}(N)]$}
        \State \quad {$\mathbf{l}^{u} = \mathbf{l}_{gt}[inds]$}
    \end{algorithmic}
    \label{alg:tla}
\end{algorithm}

As the spatial-temporal action detection tasks are always accompanied by multi-label and long-tail classification problems, pseudo-labels have a high risk of missing tags and inaccuracies, especially for the rare categories with poor classification performance. Besides, we found that the temporal constraints are strong in spatio-temporal data. Therefore, we propose temporal label assignment~(TLA) to assign classification labels for unlabeled data. Because the temporal actions are highly bounded by the temporal restrictions, we propose TLA to assign pseudo labels to detected person proposals by utilizing the neighbour annotated keyframes. The TLA procedure is detailed in Algorithm~\ref{alg:tla}. Firstly, $D_{teacher}$ generates person proposals $\mathbf{b}^{u}$ and classification scores $\mathbf{s}$ for a video clip $\mathbf{c}^u \in \mathcal{D}^u$ with unlabeled center frame. We fetch the ground-truth bounding boxes $\mathbf{b}^{left}, \mathbf{b}^{right}$ and classification labels $\mathbf{l}^{left}, \mathbf{l}^{right}$ of the neighbour annotated keyframes which are nearest to center frame of $\mathbf{c}^u$ to perform TLA. Then we assign pseudo-labels to person proposals by resorting to the help of Hungarian algorithm. Following~\cite{detreccv2020}, we consider both classification and regression factors and build the cost function between $i$-th prediction and $j$-th annotation as Eq.~\ref{eq:cost-function}.
\begin{equation}
    Cost_{ij} = \mathcal{L}_{bce}(\mathbf{s}_i, \mathbf{l}_j^{gt}) + \mathcal{L}_{L1}(\mathbf{b}_i^{u}, \mathbf{b}_j^{gt}) + \mathcal{L}_{iou}(\mathbf{b}_i^{u}, \mathbf{b}_j^{gt})
    \label{eq:cost-function}
\end{equation}
where $\mathcal{L}_{bce}, \mathcal{L}_{L1}$ and $\mathcal{L}_{iou}$ are binary cross entropy loss, smooth-L1 loss and GIoU loss. Weights of loss functions are set to 1. Then we use Hungarian algorithm~\cite{hungrian1955} to calculate the optimal label assignment policy $\hat{\pi}$ to minimize Eq.~\ref{eq:hungarian}.
\begin{equation}
    \hat{\pi} = \arg\min_{\pi \in \prod_{N}^M} \sum_i^N Cost_{i, \pi(i)}
    \label{eq:hungarian}
\end{equation}
Finally, we can use $\hat{\pi}$ to assign pseudo classification label $\mathbf{l}^{gt}_{\hat{\pi}(i)}$ to $i$-th person boxes $\mathbf{b}_i^{u}$. Each ground-truth bounding boxes could only be assigned to one person proposal. If the number of proposals $N$ is larger than the number ground-truth bounding boxes $M$, additional background objects will be added. The cost between one prediction and one background object only contains the classification part~(i.e. the binary cross entropy loss).

\section{Experiments}

In this section, we will provide the experiments settings, results, and ablations.
\subsection{Experimental Setup}

\subsubsection{Datasets}

We mainly use AVA~\cite{avacvpr2018} and JHMDB~\cite{jhmdbiccv2013} to conduct all our experiments. 

AVA~\cite{avacvpr2018} is a major dataset for benchmarking the performance of spatial-temporal action detection. It contains about 211k training clips and 57k validating video clips. The labels are annotated at 1FPS. Following standard evaluation protocol~\cite{avacvpr2018,avabaselinearxiv2018,slowfasticcv2019}, we evaluate 60 classes among the total 80 classes. We evaluate on both versions~(v2.1 and v2.2) of annotations on AVA.

JHMDB~\cite{jhmdbiccv2013} consists 21 action classes and 928 clips. JHMDB is a densely annotated dataset with per-frame annotations. Following previous works~\cite{contextrcnneccv2020,wooiccv2021}, we report the frame-level mean average precision (frame-mAP) with IoU threshold of 0.5, 
\subsubsection{Training Details} We use a server with eight 3090 GPUs to conduct all our experiments. We use PyTorch~\cite{pytorchnips2019} to implement our SE-STAD. To conduct a fair comparison, we adopt the commonly used backbone, SlowOnly and SlowFast~\cite{slowfasticcv2019} network as our backbone. We use SlowOnly ResNet50, SlowFast ResNet50 and SlowFast ResNet101 with non-local~\cite{nonlocalcvpr2018} modules to perform experiments. For the actor localization head, we use an improved version of FCOS~\cite{fcosiccv2019}, i.e., FCOS with center sampling as our actor localization head. 

\begin{table*}[t]
 \centering 
 \small
 {

  \begin{tabular}{ c|l|l|c|c|c|c|c}
  AVA & Model & Backbone & Frames & E2E & Pretrain & val mAP & GFLOPs \\
  \hline 
  \multirow{13}{*}{AVA v2.1}     
      & VAT~\cite{vatcvpr2019} & I3D & 64 & \xmark  & K400 & 25.2 & N/A\\
      & I3D~\cite{avabaselinearxiv2018} & I3D & 64 & \cmark &  K600 & 21.9 & N/A \\ 
      & Context-RCNN~\cite{contextrcnneccv2020} & R50-NL & 64 & \xmark & K400 & 28.0 & N/A \\
      & LFB~\cite{lfbcvpr2019} & R50-NL & 64 & \xmark & K400 & 25.8 & N/A \\
      & LFB~\cite{lfbcvpr2019} & R101-NL & 64 & \xmark & K400 & 27.1 & N/A \\
      \cline{2-8}
      & SlowFast~\cite{slowfasticcv2019} & \multirow{4}{*}{\begin{tabular}[x]{@{}l@{}}R50\\$8 \times 8$\end{tabular}} & 32& \xmark & K400 & 24.7 & 97.5+406.5\\
      & WOO$^*$~\cite{wooiccv2021} & & 32& \xmark & K400 & 25.2 & 141.6\\
      & SE-STAD &  & 32 & \cmark & K400 & 25.0 &  111.3\\
      & SE-STAD + TLA &  & 32 & \cmark & K400 & \textbf{26.5} & 111.3\\
      \cline{2-8}
      & SlowFast~\cite{slowfasticcv2019} & \multirow{4}{*}{\begin{tabular}[x]{@{}l@{}}R101-NL\\$8 \times 8$\end{tabular}} & 32 & \xmark & K600 & 27.3 & 151.5+406.5\\
      & WOO$^*$~\cite{wooiccv2021} &  & 32 & \cmark & K600 & 28.0 & 245.8 \\
      & SE-STAD &  & 32 & \cmark & K600 & 27.7 &  165.2\\
      & SE-STAD + TLA &  & 32 & \cmark & K600 & \textbf{28.8} & 165.2\\
      \cline{2-8} 
      & SE-STAD + TLA$^*$ & \multirow{2}{*}{\begin{tabular}[x]{@{}l@{}}R101\\$8 \times 8$\end{tabular}} & 32 & \checkmark & K700 & \textbf{31.8} & 192.7 \\
      & TubeR~\cite{tubercvpr2022} & & 32  & \checkmark &  K700 & 31.6 & 240 \\
    \hline 
    
   \multirow{12}{*}{AVA v2.2} 
    & SlowOnly~\cite{slowfasticcv2019} & \multirow{4}{*}{\begin{tabular}[x]{@{}l@{}}R50\\$4 \times 16$\end{tabular}} & 4 & \xmark & K400 & 20.3 & 41.8+406.5\\
   & WOO$^*$~\cite{wooiccv2021} & & 4 & \cmark & K400 & 21.3 & 68.0 \\
   & SE-STAD & & 4 & \cmark & K400 & 21.5 & 55.5\\
   & SE-STAD + TLA & & 4 & \cmark & K400 & \textbf{22.0} & 55.5\\
   \cline{2-8} 
   & SlowFast~\cite{slowfasticcv2019} & \multirow{4}{*}{\begin{tabular}[x]{@{}l@{}}R50\\$8 \times 8$\end{tabular}} & 32  & \xmark & K400 & 24.7 & 97.5+406.5\\
   & WOO$^*$~\cite{wooiccv2021} &  & 32 & \cmark & K400 & 25.4 & 147.5 \\ 
   & SE-STAD &  & 32 & \cmark & K400 & 25.5 &  111.3\\
   & SE-STAD + TLA &  & 32 & \cmark & K400 & \textbf{26.9} & 111.3\\

   \cline{2-8} 
   & SlowFast~\cite{slowfasticcv2019} & \multirow{4}{*}{\begin{tabular}[x]{@{}l@{}}R101-NL\\$8 \times 8$\end{tabular}} & 32 & \xmark & K600 & 27.4 & 151.5+406.5\\
   & WOO$^*$~\cite{wooiccv2021} &  & 32 & \cmark & K600 & 28.3 & 251.7 \\
   & SE-STAD &  & 32 & \cmark & K600 & 28.5 & 165.2\\
   & SE-STAD + TLA &  & 32 & \cmark & K600 & \textbf{29.3} & 165.2\\

   \hline
  \end{tabular}
   \caption{\label{tab:result_ava} \textbf{Results on AVA dataset.} We report the FLOPs of action classification network plus the FLOPs of person detector for proposal-based methods. We calculate the FLOPs of person detector according to the official configure file provided by~\cite{lfbcvpr2019}. $^*$ means the method reports the performance by testing with 320 resolution.}
   \vspace{-1em}
 }

\end{table*}

Following previous works~\cite{avabaselinearxiv2018,slowfasticcv2019}, we use SGD with momentum as our optimizer. The hyperparameters are listed as follows: The batch size is 48 with 8 GPUs (6 clips per GPU) for the burn-in~(baseline) stage, and 96 in the semi-supervised action detection~(SSAD) stage. the ratio of labeled/unlabeled data is 1:1 in the SSAD stage. We use an initial learning rate of 0.075 and the cosine decay schedule. We train the model with 20000 iterations~(around 5 epochs) in the burn-in stage, and we train the model with 40000 iterations~(around 10 epochs) in the SSAD stage. For models without SSAD, we train the model with 40000 iterations. Longer training schedules~(60000 or 80000 iterations) will decrease the performance by around 0.3\% mAP when adopting SlowFast R50 backbone. The backbone is initialized with the pre-trained weights on Kinetics-400 or Kinetics-600~\cite{kineticscvpr2017}. The actor localization head uses the initialization schedule in the original FCOS~\cite{fcosiccv2019} paper. For other layers, we initialize the layer with Xavier~\cite{xavieraistats2010}. We perform random scaling to the video clip input, we random resize the shortest edge to $[256, 320]$. and then we random crop a $256 \times 256$ video clip to feed into the model. 

For the actor localization head, we will use a post-processing step with 0.3 scoring threshold and maximum number of 100 proposals during training. We \texttt{do not} perform non-maximum suppression~(NMS) for actor localization head in the training stage. Then those proposals will be fed into the action classification head. The loss is showed in Sec.~3. The loss weight for $\mathcal{L}_{al}$ is 1 and 10 for $\mathcal{L}_{ac}$. Generated proposals which have at least 50\% intersection-over-union~(IoU) with the ground-truth boxes will be treated as positive proposals in the action classification stage, otherwise those proposals will be ignored.
\vspace{-1.3em}
\subsubsection{Testing Details} 
\vspace{-0.5em}
Our inference steps are somehow simple. With an input video clip, we will first resize the shortest edge to 256, then directly feed into the model. We will use a post-processing step with 0.4 scoring threshold and NMS step with IoU threshold 0.3 to get the testing proposals. Then those proposals will be fed into the action classification head. We set the final action threshold as 0.002 and limit the maximum output of actors to 10 per image. During inference, we always use a single view instead of applying multi-scale testing to our models.

\subsection{Results on AVA}

In this section, we will provide results and analyses of results on AVA. The results are listed in Table~\ref{tab:result_ava}. From the table, we can have the following observations:

\squishlist
\item The extra person detector will bring a huge computational burden to the spatial-temporal action detection model. The FLOPs of the person detector is 406.5G FLOPs. which is around 7 times larger than the FLOPs of SlowOnly R50~($4 \times 16$), and more than 2 times larger than the heaviest backbone SlowFast R101-NL~($8 \times 8$). The large FLOPs come from the high input resolution in the person detector. The high input resolution along with the high FLOPs makes proposal-based methods hard to apply in real-world scenarios.

\item With a simply added component, i.e., FCOS head, our model can have roughly comparable or better performance than proposal-based methods, even than the recent WOO~\cite{wooiccv2021} and we do not use extra SSAD techniques. This is quite encouraging because we have around 70$\sim$90\% FLOPs drop with proposal-based SlowFast, and we have around 20$\sim$35\% FLOPs drop with WOO~\cite{wooiccv2021}. This shows the effectiveness of our simple models. We will dive into the model details part to figure out what makes the simple model work so well.

\item With the extra SSAD techniques~(the semi-supervised learning stage and the temporal labeling assignment), our model can have an extra performance boost with no extra modules and no computational cost. For example, SlowFast R50 can have an extra 1.4\% mAP performance boost on AVA v2.2 and 1.5\% mAP boost on AVA v2.1. Similar performance gaps are observed on SlowFast R101. However, SSAD stage can only have a 0.5\% performance gain on SlowOnly R50. We conjecture that it may be due to the input capacity. SlowOnly R50 only has 4 frames as input. The low number of input frames prevents SlowOnly R50 to have better performance. We can achieve 31.8 mAP on AVA v2.1, which is 0.2 mAP higher than TubeR~\cite{tubercvpr2022}, and our model has 20\% fewer FLOPs than TubeR. TubeR uses extra encoder-decoder structure with Transformers to perform end-to-end STAD. Our SE-STAD, have comparable or better performance and simple design.
\squishend

\subsection{Results on JHMDB}

To verify the effectiveness of SE-STAD, we further evaluate our model on JHMDB~\cite{jhmdbiccv2013}. Since JHMDB is densely-annotated, we directly apply the basic SE-STAD model. The results are in Table~\ref{tab:result_jhmdb}. From the table, we can observe that: SE-STAD models can achieve 82.5\% mAP with SlowFast R101 8x8 backbone, which is 2.0\% higher than WOO~\cite{wooiccv2021}. Even with a weaker backbone, SE-STAD can still achieve 80.7\% mAP, which is still 0.2\% higher than WOO. These results show the effectiveness of SE-STAD.
\begin{table}[t]
   \centering
   \scriptsize
   \setlength{\tabcolsep}{2.5pt}
   \begin{tabular}{|c|c|c|}
   Method  & Backbone    & JHMDB mAP  \\ \hline
   Context-Aware RCNN~\cite{contextrcnneccv2020} & I3D R50-NL 8x8 & 79.2 \\
   WOO~\cite{wooiccv2021}     & SlowFast R101-NL 8x8 & 80.5 \\ \hline
   SE-STAD & SlowFast R50 8x8 & 80.7 \\
   SE-STAD & SlowFast R101-NL 8x8 & \textbf{82.5} \\ 
   \end{tabular}
       \caption{\label{tab:result_jhmdb} \textbf{Results on JHMDB dataset.}}
   \vspace{-2em}

   \end{table}
   
\subsection{Ablation Study}

In this section, we will provide ablations of our model, including the head choice for actor localization, loss coefficients, input resolutions and the methods to train the  classification head. In this section, unless specified, all experiments use SlowFast R50~($8 \times 8$) as the backbone network. 
\subsubsection{Head Choices for Actor Localization}
In this section, we vary the head of actor localization for SE-STAD. We try different heads, including the popular anchor-based heads: RPN + RCNN~\cite{fasterrcnnnips2015}, RetinaNet~\cite{retinaneticcv2017} and GFocalV2~\cite{gfocalv2cvpr2021}. The ablation results are in Table~\ref{ablation1}. From Table~\ref{ablation1} we can observe that, Anchor-based heads perform significantly worse than anchor-free heads, i.e., FCOS~\cite{fcosiccv2019}. Two-stage RPN+RCNN~\cite{fasterrcnnnips2015} and RetinaNet~\cite{retinaneticcv2017} have a large performance drop. Even the most recent GFocalV2 head~(anchor-based version) will have a 1.8\% \textit{m}AP gap with the FCOS head. Besides, tricks on FCOS will improve around 0.6\% mAP, and the original FCOS head will still achieve a 24.9\% mAP.
It may be due to the low input resolution and pre-defined anchor shape. Moreover, with the simple FCOS head, our model performs slightly better than WOO~\cite{wooiccv2021}. WOO has an extra attention module. In contrast, our SE-STAD keeps a simple architectural design and has a good performance.

\begin{table}[t]
        \centering
        \resizebox{0.7\linewidth}{!}{
        \begin{tabular}{l|c|c}
        Actor Heads  & Type & \textit{m}AP          \\
        \hline
        RPN+RCNN~\cite{fasterrcnnnips2015}    & Anchor-Based & 21.0  \\
        RetinaNet~\cite{retinaneticcv2017} & Anchor-Based & 19.7 \\
        GFocalV2~\cite{gfocalv2cvpr2021} & Anchor-Based & 23.7      \\
        FCOS-~\cite{fcosiccv2019} & Anchor-Free & 24.9 \\
        WOO*~\cite{wooiccv2021} & Anchor-Free & 25.4 \\
        FCOS~\cite{fcosiccv2019} & Anchor-Free & 25.5 \\
        \end{tabular}}
        \caption{\label{ablation1}\textbf{Ablation study on different heads for actor localization}. We try different heads with SlowFast R50 backbone. We apply the anchor-based version of GFocalV2~\cite{gfocalv2cvpr2021}. FCOS- is the original FCOS~\cite{fcosiccv2019} version without tricks. We do not use self/semi-training for all methods. WOO~\cite{wooiccv2021} is listed only for comparison. }
\vspace{-1.5em}
\end{table}

\subsubsection{Different Strategies to Train Action Classification}

The action classification part is the other important part for SE-STAD. We will use different strategies to train and test our models to ablate our FCOS head. We vary the input of action classification head between training and testing, and verify the performance of our model.

The results are in Table~\ref{ablation2}. We can find that:
\squishlist
\item When testing with pre-extracted proposals, our model can have better performance than FCOS generated boxes. It is not surprising because we are performing actor localization with low-resolution inputs. 
\item However, our model still performs better than proposal-based methods. Also, our model trained with sparse inputs~(GT, Sparse FCOS outputs) performs worse than dense inputs with a more than 1\% mAP gap. This result shows that we should use dense inputs to boost the classification performance. This can be an explanation of why WOO performs badly with Sparse-RCNN~\cite{sparsercnncvpr2021}.  

\squishend

\begin{table}[t]
        \centering
        \resizebox{0.7\linewidth}{!}{

        \begin{tabular}{l|c|c}
        Training input  & Testing input  & \textit{m}AP          \\
        \hline
        Proposal & Proposal & 24.7 \\
        GT Only  & Proposal & 24.5  \\
        GT Only & FCOS Output & 23.7 \\
        FCOS Output~(Sparse) & FCOS Output & 24.3 \\
        FCOS Output~(Dense) & FCOS Output & 25.5 \\
        FCOS Output~(Dense) & Proposal & 26.2 \\
        \end{tabular}}
        \caption{\label{ablation2}\textbf{Ablation study on different training inputs for action classification}. We try different inputs for both training and testing of our models. For ``GT Only'', we only feed the ground-truth boxes into the action classification head. For ``FCOS Output~(Sparse)'', we perform NMS to FCOS generated proposals in the training stage. For ``FCOS Output~(Dense)'', we do not perform NMS in the training stage.}
\end{table}

\vspace{-1.5em}

\subsubsection{Ablations on Loss coefficients}

\vspace{-0.5em}

As stated in Sec.~3, we introduce $\lambda_{cls}$ to balance the actor localization and action classification loss. We also introduce $\lambda_{unsup}$ to balance labeled and unlabeled losses. Here, we make ablation experiments to show the robustness of each coefficient.  Results in Table~\ref{ablation2} support the robustness of $\lambda_{cls}$ and $\lambda_{unsup}$. $\lambda_{cls} = 10$ achieves best performance. Besides, when utilizing the unlabeled data, ablation experiments show that it's better to set the ratio of loss weight between the labeled part and unlabeled part to 2:1.

\vspace{-0.6em}
\subsubsection{Computational Efficiency}

\begin{table}[t]
        \centering
        \resizebox{0.7\linewidth}{!}{
        \begin{tabular}{c|c|c|c}
        Method       & $\lambda_{cls}$ & $\lambda_{unsup}$ & val mAP \\ \hline
        SE-STAD      & 1               & -                &      25.2    \\
        SE-STAD     & 10               & -                &    25.5     \\
        SE-STAD     & 20              & -                &    24.7   \\
        SE-STAD+TLA & 10              & 0.2              &    26.8   \\
        SE-STAD+TLA & 10              & 0.5              &    26.9    \\
        SE-STAD+TLA & 10              & 1.0              &    26.5    
        \end{tabular} }
        \caption{\label{ablation3}\textbf{Ablation study on $\lambda_{unsup}$ and $\lambda_{cls}$}. We study the effect of $\lambda_{unsup}$ and $\lambda_{cls}$ to verify the robustness of SE-STAD.}
        \vspace{-0.4cm}
\end{table}

\vspace{-0.6em}

In this section, we will show the computational efficiency of our model under different input resolutions.

We vary the input resolution for our model during testing. The results are in Table~\ref{ablation4}. We can observe that with the default 256 input resolution, we perform slightly better than WOO~\cite{wooiccv2021} and proposal-based SlowFast~\cite{slowfasticcv2019}. When we use a larger input resolution, i.e., 320, we can get a 0.6\% performance boost and slightly higher FLOPs than WOO~\cite{wooiccv2021} but much lower than proposal-based SlowFast. 

\vspace{-0.7em}
\subsubsection{Ablations on Pseudo Label Generation} 
\vspace{-0.5em}

For the semi-supervised action detection part, pseudo label generation is the critical part for this part. It's a natural way to perform pseudo label generation by predicting the classification labels directly. However, as stated before, multi-label and long-tail classification problems make the SSAD part hard. In order to show the excellence of TLA, we explore different strategies to generate pseudo labels:

\begin{table}[t]
        \centering

        \resizebox{0.5\linewidth}{!}{
        \begin{tabular}{l|c}
        Pseudo Label  & Performance         \\
        \hline
        None & 25.5 \\
        Interpolation & 24.8 \\
        EMA & 26.0 \\
        Hard Threshold & 26.0 \\
        Per Class Threshold & 26.2 \\
        TLA & 26.9 \\
        \end{tabular}}
        \caption{\label{ablation5}\textbf{Ablation study on different strategies to generate pseudo labels for the SSAD stage on AVA v2.2}. We also report a strong baseline: training on annotated frames with EMA.}
    \vspace{-1em}
\end{table}

\begin{table}[t]
\centering

\resizebox{0.75\linewidth}{!}{
\begin{tabular}{l|c|c|c|c|}

Models & Backbone & Input Res & Performance & FLOPs         \\
\hline
SlowFast &\multirow{4}{*}{R50, $8 \times 8$} & 256 & 24.7 & 97.5+406.5 \\ 
WOO & & 320 & 25.4 & 147.5 \\
Ours & & 256 & 25.5 & 111.3\\ 
Ours & & 320 & 26.1 & 173.8\\
\end{tabular}}
\caption{\label{ablation4}\textbf{Ablation study on input resolutions on AVA v2.2}. We use the square input, e.g., $256 \times 256$ to calculate the FLOPs for all our models. For WOO, we directly report the result from~\cite{wooiccv2021}.}
\vspace{-1em}
\end{table}

\squishlist
\item Hard Threshold. We apply a hard threshold for all classes to filter generated pseudo boxes.
\item Per Class Threshold: We apply an independent threshold for each class to filter pseudo boxes. Thresholds are calculated from the model on the training set.
\item TLA: The method we proposed in Sec.~3.
\squishend

For the former two strategies, we apply a temporal labeling restriction additionally to boost the performance: we remove classes that are not in the union set of surrounding annotated frames. Besides those semi-supervised techniques, we also try a strong but simple baseline: We discard the teacher model, and the model is learned with EMA on the annotated subset. The results are in Table~\ref{ablation5}. 

From the table, we can observe: the commonly used hard threshold strategy does not work on the AVA dataset if we consider the influence of EMA. Even if we consider the multi-label and long-tail problems in the dataset, and have a stronger per class threshold baseline, it still has a minor performance gain over the baseline model. In contrast, our TLA along with SSAD has a better performance, which shows the effectiveness of TLA.

\section{Conclusion}
In this paper, we presented SE-STAD, an end-to-end method for spatial-temporal action detection. SE-STAD has a simple design and small computational burden, yet achieves good results across the major spatial-temporal action detection dataset. The performance gain comes from two parts: one is the powerful anchor-free detector head. The other is the proposed novel semi-supervised training schema along with the label assignment strategy. We hope that our model, notwithstanding its simplicity, can en-light the broader problem of video understanding. We will continue to explore the multi-label and long-tailed problems that existed in spatial-temporal action detection.

\clearpage
\appendix

\section{Appendix}

This appendix provides further details of the main paper. We present extra experiments and visualizations. (1) Our SE-STAD can cooperate with stronger action classification heads; (2) The ablation of feature pyramid architecture; (3) Additional visualizations of TLA.

\section{Cooperating with stronger action classification heads}
In this paper, we dive deep into building an end-to-end spatial-temporal action detector with minimum efforts. Contrary to~\cite{wooiccv2021,point3dbmvc2021} which highly rely on extra attention mechanisms, no additional attention module is introduced in our proposed SE-STAD. In order to show that our SE-STAD could also benefit from attention modules, we adopt a simple ACRN head~\cite{acrneccv2018} as the action classification head. Table~\ref{tab:attention} shows the experiment results. We could find that the basic SE-STAD could get 1.5\% gains from the introduced simple ACRN head. Such results demonstrate that we could further boost the performance with other fancy attention modules, such as~\cite{acrneccv2018,acarcvpr2021}.

\begin{table}[h]
\centering
\begin{tabular}{c|c|c}
Method  & Attention Module & val mAP  \\ \hline
SE-STAD & None             & 25.5 \\
SE-STAD & ACRN~\cite{acrneccv2018}             & 27.0
\end{tabular}
\caption{\label{tab:attention}\textbf{Ablation study on classification head}. We try to use ACRN~\cite{acrneccv2018} head as the classification head. Experiments are performed with SlowFast R50 backbone.}
\end{table}

\section{Cooperating with other backbones}
For fair comparisons, we only perform experiments with SlowFast backbone in the main paper. To show the generalizability of our model, we further perform experiments with I3D~\cite{kineticscvpr2017} network. As shown in Table~\ref{tab:backbone}, we could reach 23.0 mAP with the I3D backbone which is still far better than AVA baseline~\cite{avacvpr2018} and result obtained with more pretraining data and heavy head~\cite{avabaselinearxiv2018}. 

\begin{table}[h]
\centering
\resizebox{0.98\linewidth}{!}{
\begin{tabular}{c|c|c|c}
Method  & Backbone & Pretrain & val mAP  \\ \hline
AVA baseline~\cite{avacvpr2018} & I3D~\cite{acrneccv2018}      &    K400   & 15.8 \\
Better baseline$^*$~\cite{avabaselinearxiv2018} & I3D~\cite{acrneccv2018}     &    K600        & 21.9 \\
SE-STAD & I3D~\cite{acrneccv2018}             &  K400  &    23.0 \\
SE-STAD & SlowFast R50~\cite{slowfasticcv2019}             &  K400  &    25.0
\end{tabular}
}
\caption{\label{tab:backbone}\textbf{Ablation study on different backbones}. We try to use I3D as the backbone of SE-STAD. No additional attention module is introduced. $^*$ means the ``Better baseline'' method use heave I3D blocks to perform action classification.}
\end{table}

\section{Ablation of feature pyramid architecture}
As shown in Fig. 2 in this paper, we build the feature pyramid~(P3-P5) on top of features of keyframes from Res3 and Res4 layers. 
We adopt such an architecture after considering the balance of computation complexity and performance. 
We do extra experiments about adopting different feature pyramid architectures to perform actor localization.
Experiment results are shown in Table~\ref{tab:fpn}. These results show that the architecture of the feature pyramid does not play the most essential role in SE-STAD. Although building P2-5 on Res2-4 could boost the performance from 25.5 to 26.0, it will introduce around 40\% computation additionally. Hence, we select to build P3-P5 based on Res3 and Res4 to make a trade-off between the detector performance and computation complexity.
 

\begin{table}[h]
\centering
\begin{tabular}{c|c|c|c|c}
Method  & Source & Target & val mAP & GFLOPs \\ \hline
SE-STAD & Res3-4 & P3-5   & 25.5    & 111.3  \\
SE-STAD & Res3-5 & P3-5   & 25.2    & 113.4  \\
SE-STAD & Res2-4 & P2-5   & 26.0    & 152.8 
\end{tabular}
\caption{\label{tab:fpn}\textbf{Ablation study on the architecture of feature pyramid}. We try different feature pyramid architectures with SlowFast R50 backbone.}
\end{table}

\section{FLOPs Analysis}

\begin{figure}[h]
    \centering
    \includegraphics[width=0.45\textwidth]{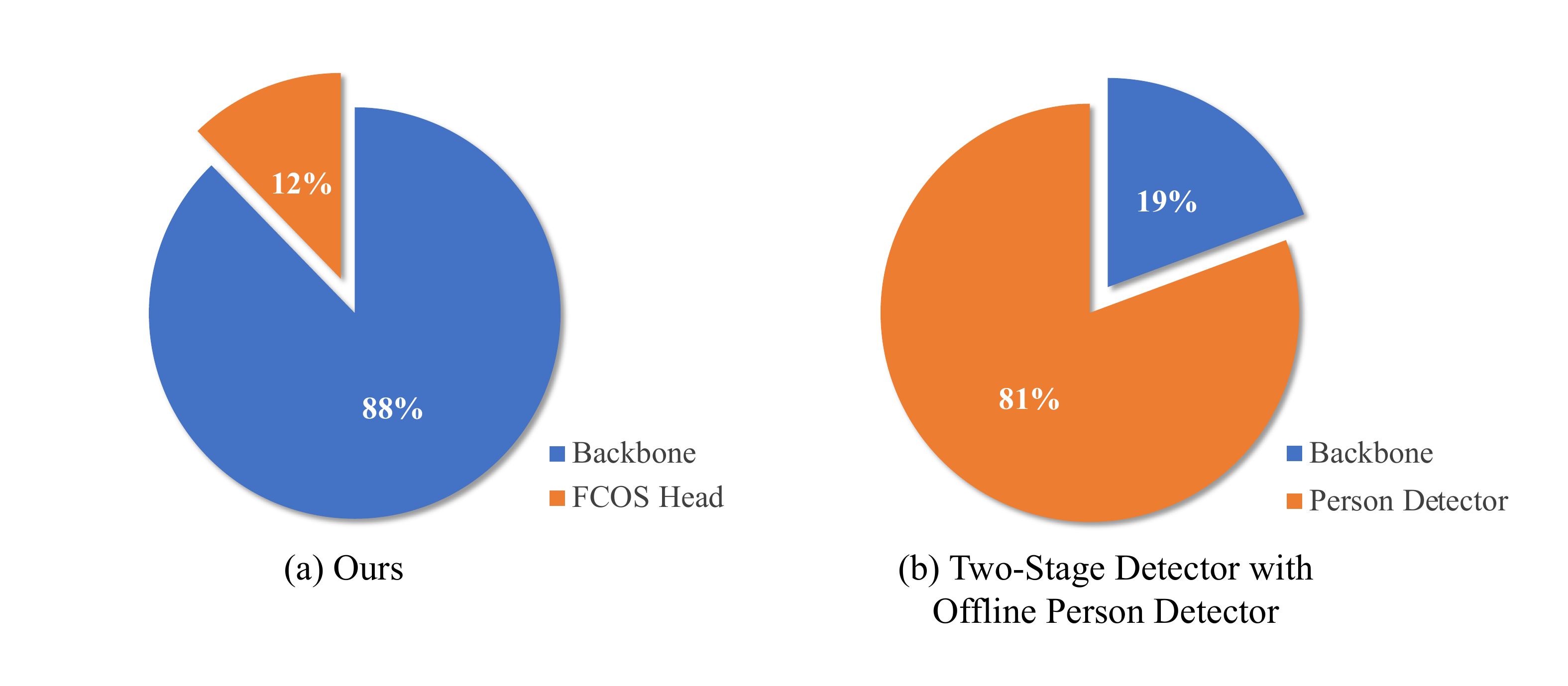}
    \caption{FLOPs pie chart with different parts in the spatial-temporal action detector. The left figure is our SE-STAD and the right figure is the proposal-based SlowFast. FLOPs of action classification head is ignored as the computation complexity of this part is too small. Both methods use the same SlowFast R50 backbone.}
    \label{fig:computation}
\end{figure}

We analyse the component of different spatial-temporal action detectors, including proposal-based methods and our SE-STAD. The analysis results are in Fig.~\ref{fig:computation}. From Fig.~\ref{fig:computation}, we can easily observe that: FCOS head only occupies around 12\% percent of the whole SE-STAD, and backbone action classification network has the majority computational complexity in SE-STAD. In contrast, the majority computational burden of two-stage detectors lies in the person detector part, which is redundant. 

\begin{figure*}[h!]
    \centering
    \includegraphics[width=1\textwidth]{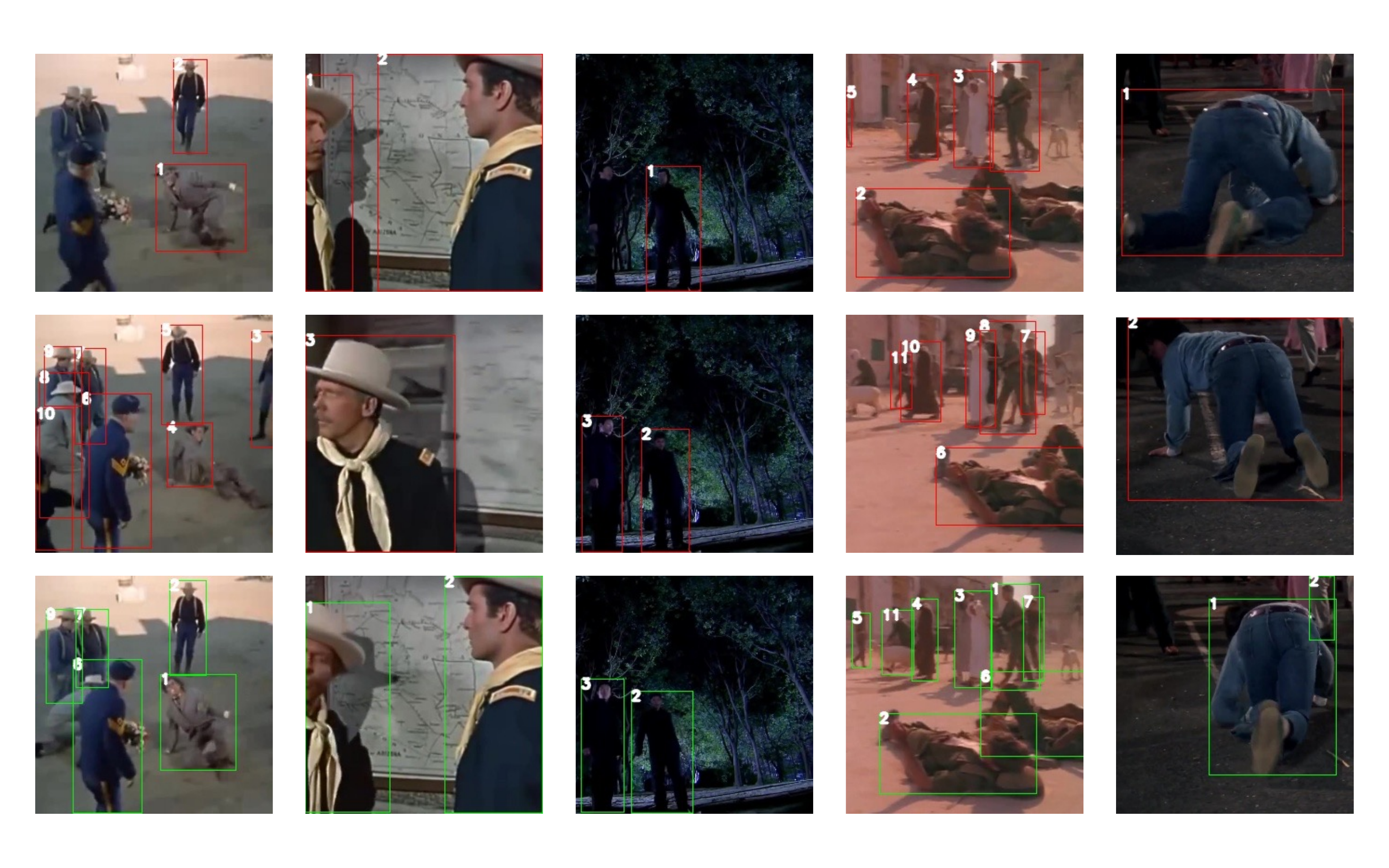}
    \caption{\textbf{Visualization of results generated by TLA.} Top row: the former neighbour keyframe with ground-truth annotations. Medium row: the later neighbour keyframe with ground-truth annotations. Bottom row: the frame labeled by TLA. Last column: a failure case. The figure is best viewed when zoomed in. Numbers of person proposals indicate the entity ids in keyframes with ground-truth annotations and assigned entity ids in frames labeled by TLA.}
    \label{fig:vis1}
\end{figure*}

\section{Visualization of TLA}
In Sec.~3, we propose a novel labeling strategy, i.e., the temporal label assignment~(TLA),  to better utilize every possible piece of information in sparse annotated spatial-temporal action detection datasets. TLA which utilizes the temporal restriction could provide more clear temporal action boundaries and fine-grained information to the detector. In Fig.~\ref{fig:vis1}, we visualize some results produced by TLA. The visualization results illustrate that TLA produces relatively reliable pseudo labels successfully on unlabeled frames. In addition, we find that there are often missing labels in provided ground-truth annotations which deteriorate the performance of the spatial-temporal action detector. Whereas, thanks to the promising actor localization ability of SE-STAD, TLA could detect most of the actors and assign dependable labels.

\clearpage 
{\small
\bibliographystyle{ieee_fullname}
\bibliography{egbib}

\begin{thebibliography}{10}\itemsep=-1pt

\bibitem{viviticcv2021}
Anurag Arnab, Mostafa Dehghani, Georg Heigold, Chen Sun, Mario Lu{\v{c}}i{\'c},
  and Cordelia Schmid.
\newblock {ViViT}: A video vision transformer.
\newblock In {\em Int. Conf. Comput. Vis.}, pages 6836--6846, 2021.

\bibitem{mixmatchnips2019}
David Berthelot, Nicholas Carlini, Ian Goodfellow, Nicolas Papernot, Avital
  Oliver, and Colin~A Raffel.
\newblock Mixmatch: A holistic approach to semi-supervised learning.
\newblock {\em Adv. Neural Inform. Process. Syst.}, 32, 2019.

\bibitem{detreccv2020}
Nicolas Carion, Francisco Massa, Gabriel Synnaeve, Nicolas Usunier, Alexander
  Kirillov, and Sergey Zagoruyko.
\newblock End-to-end object detection with transformers.
\newblock In {\em Eur. Conf. Comput. Vis.}, pages 213--229, 2020.

\bibitem{kineticscvpr2017}
Joao Carreira and Andrew Zisserman.
\newblock Quo vadis, action recognition? a new model and the kinetics dataset.
\newblock In {\em IEEE Conf. Comput. Vis. Pattern Recog.}, pages 6299--6308,
  2017.

\bibitem{wooiccv2021}
Shoufa Chen, Peize Sun, Enze Xie, Chongjian Ge, Jiannan Wu, Lan Ma, Jiajun
  Shen, and Ping Luo.
\newblock Watch only once: An end-to-end video action detection framework.
\newblock In {\em Int. Conf. Comput. Vis.}, pages 8178--8187, 2021.

\bibitem{mviticcv2021}
Haoqi Fan, Bo Xiong, Karttikeya Mangalam, Yanghao Li, Zhicheng Yan, Jitendra
  Malik, and Christoph Feichtenhofer.
\newblock Multiscale vision transformers.
\newblock In {\em Int. Conf. Comput. Vis.}, pages 6824--6835, 2021.

\bibitem{slowfasticcv2019}
Christoph Feichtenhofer, Haoqi Fan, Jitendra Malik, and Kaiming He.
\newblock {SlowFast} networks for video recognition.
\newblock In {\em Int. Conf. Comput. Vis.}, pages 6202--6211, 2019.

\bibitem{avabaselinearxiv2018}
Rohit Girdhar, Joao Carreira, Carl Doersch, and Andrew Zisserman.
\newblock A better baseline for {AVA}.
\newblock {\em arXiv preprint arXiv:1807.10066}, 2018.

\bibitem{vatcvpr2019}
Rohit Girdhar, Joao Carreira, Carl Doersch, and Andrew Zisserman.
\newblock Video action transformer network.
\newblock In {\em IEEE Conf. Comput. Vis. Pattern Recog.}, pages 244--253,
  2019.

\bibitem{fastrcnniccv2015}
Ross Girshick.
\newblock Fast {R-CNN}.
\newblock In {\em Int. Conf. Comput. Vis.}, pages 1440--1448, 2015.

\bibitem{rcnncvpr2014}
Ross Girshick, Jeff Donahue, Trevor Darrell, and Jitendra Malik.
\newblock Rich feature hierarchies for accurate object detection and semantic
  segmentation.
\newblock In {\em IEEE Conf. Comput. Vis. Pattern Recog.}, pages 580--587,
  2014.

\bibitem{xavieraistats2010}
Xavier Glorot and Yoshua Bengio.
\newblock Understanding the difficulty of training deep feedforward neural
  networks.
\newblock In {\em Int. Conf. Arti. Intell. Stat.}, pages 249--256, 2010.

\bibitem{avacvpr2018}
Chunhui Gu, Chen Sun, David~A Ross, Carl Vondrick, Caroline Pantofaru, Yeqing
  Li, Sudheendra Vijayanarasimhan, George Toderici, Susanna Ricco, Rahul
  Sukthankar, et~al.
\newblock {AVA}: A video dataset of spatio-temporally localized atomic visual
  actions.
\newblock In {\em IEEE Conf. Comput. Vis. Pattern Recog.}, pages 6047--6056,
  2018.

\bibitem{csdnips2019}
Jisoo Jeong, Seungeui Lee, Jeesoo Kim, and Nojun Kwak.
\newblock Consistency-based semi-supervised learning for object detection.
\newblock {\em Adv. Neural Inform. Process. Syst.}, 32, 2019.

\bibitem{jhmdbiccv2013}
H. Jhuang, J. Gall, S. Zuffi, C. Schmid, and M.~J. Black.
\newblock Towards understanding action recognition.
\newblock In {\em Int. Conf. Comput. Vis.}, pages 3192--3199, 2013.

\bibitem{yowoarxiv2019}
Okan K{\"o}p{\"u}kl{\"u}, Xiangyu Wei, and Gerhard Rigoll.
\newblock You only watch once: A unified {CNN} architecture for real-time
  spatiotemporal action localization.
\newblock {\em arXiv preprint arXiv:1911.06644}, 2019.

\bibitem{hungrian1955}
Harold~W Kuhn.
\newblock The {Hungarian} method for the assignment problem.
\newblock {\em Naval research logistics quarterly}, 2(1-2):83--97, 1955.

\bibitem{pseudolabelicmlworkshop2013}
Dong-Hyun Lee et~al.
\newblock Pseudo-label: The simple and efficient semi-supervised learning
  method for deep neural networks.
\newblock In {\em Int. Conf. Mach. Learn. Workshops}, page 896, 2013.

\bibitem{avakineticsarxiv2020}
Ang Li, Meghana Thotakuri, David~A Ross, Jo{\~a}o Carreira, Alexander
  Vostrikov, and Andrew Zisserman.
\newblock The {AVA}-{Kinetics} localized human actions video dataset.
\newblock {\em arXiv preprint arXiv:2005.00214}, 2020.

\bibitem{rtpeccv2018}
Dong Li, Zhaofan Qiu, Qi Dai, Ting Yao, and Tao Mei.
\newblock Recurrent tubelet proposal and recognition networks for action
  detection.
\newblock In {\em ECCV}, pages 303--318, 2018.

\bibitem{gfocalv2cvpr2021}
Xiang Li, Wenhai Wang, Xiaolin Hu, Jun Li, Jinhui Tang, and Jian Yang.
\newblock Generalized focal loss v2: Learning reliable localization quality
  estimation for dense object detection.
\newblock In {\em IEEE Conf. Comput. Vis. Pattern Recog.}, pages 11632--11641,
  2021.

\bibitem{multisportsiccv2021}
Yixuan Li, Lei Chen, Runyu He, Zhenzhi Wang, Gangshan Wu, and Limin Wang.
\newblock {MultiSports}: A multi-person video dataset of spatio-temporally
  localized sports actions.
\newblock In {\em Int. Conf. Comput. Vis.}, pages 13536--13545, 2021.

\bibitem{retinaneticcv2017}
Tsung-Yi Lin, Priya Goyal, Ross Girshick, Kaiming He, and Piotr Doll{\'a}r.
\newblock Focal loss for dense object detection.
\newblock In {\em Int. Conf. Comput. Vis.}, pages 2980--2988, 2017.

\bibitem{cocoeccv2014}
Tsung-Yi Lin, Michael Maire, Serge Belongie, James Hays, Pietro Perona, Deva
  Ramanan, Piotr Doll{\'a}r, and C~Lawrence Zitnick.
\newblock Microsoft {COCO}: Common objects in context.
\newblock In {\em Eur. Conf. Comput. Vis.}, volume 8693 of {\em LNCS}, pages
  740--755, 2014.

\bibitem{ssdeccv2016}
Wei Liu, Dragomir Anguelov, Dumitru Erhan, Christian Szegedy, Scott Reed,
  Cheng-Yang Fu, and Alexander~C Berg.
\newblock {SSD}: Single shot multibox detector.
\newblock In {\em Eur. Conf. Comput. Vis.}, pages 21--37, 2016.

\bibitem{unbiasediclr2021}
Yen-Cheng Liu, Chih-Yao Ma, Zijian He, Chia-Wen Kuo, Kan Chen, Peizhao Zhang,
  Bichen Wu, Zsolt Kira, and Peter Vajda.
\newblock Unbiased teacher for semi-supervised object detection.
\newblock In {\em Int. Conf. Learn. Represent.}, pages 1--13, 2021.

\bibitem{videoswimarxiv2021}
Ze Liu, Jia Ning, Yue Cao, Yixuan Wei, Zheng Zhang, Stephen Lin, and Han Hu.
\newblock Video swin transformer.
\newblock {\em arXiv preprint arXiv:2106.13230}, 2021.

\bibitem{point3dbmvc2021}
Shentong Mo, Jingfei Xia, Xiaoqing Tan, and Bhiksha Raj.
\newblock {Point3D}: tracking actions as moving points with 3d cnns.
\newblock In {\em Brit. Mach. Vis. Conf.}, pages 1--14, 2021.

\bibitem{acarcvpr2021}
Junting Pan, Siyu Chen, Mike~Zheng Shou, Yu Liu, Jing Shao, and Hongsheng Li.
\newblock Actor-context-actor relation network for spatio-temporal action
  localization.
\newblock In {\em IEEE Conf. Comput. Vis. Pattern Recog.}, pages 464--474,
  2021.

\bibitem{pytorchnips2019}
Adam Paszke, Sam Gross, Francisco Massa, Adam Lerer, James Bradbury, Gregory
  Chanan, Trevor Killeen, Zeming Lin, Natalia Gimelshein, Luca Antiga, et~al.
\newblock {PyTorch}: An imperative style, high-performance deep learning
  library.
\newblock In {\em Adv. Neural Inform. Process. Syst.}, volume~32, pages 1--12,
  2019.

\bibitem{pimodelnips2015}
Antti Rasmus, Mathias Berglund, Mikko Honkala, Harri Valpola, and Tapani Raiko.
\newblock Semi-supervised learning with ladder networks.
\newblock {\em Adv. Neural Inform. Process. Syst.}, 28, 2015.

\bibitem{yolocvpr2016}
Joseph Redmon, Santosh Divvala, Ross Girshick, and Ali Farhadi.
\newblock You only look once: Unified, real-time object detection.
\newblock In {\em IEEE Conf. Comput. Vis. Pattern Recog.}, pages 779--788,
  2016.

\bibitem{fasterrcnnnips2015}
Shaoqing Ren, Kaiming He, Ross Girshick, and Jian Sun.
\newblock Faster {R-CNN}: Towards real-time object detection with region
  proposal networks.
\newblock {\em Adv. Neural Inform. Process. Syst.}, 28:1--9, 2015.

\bibitem{gioulosscvpr2019}
Hamid Rezatofighi, Nathan Tsoi, JunYoung Gwak, Amir Sadeghian, Ian Reid, and
  Silvio Savarese.
\newblock Generalized intersection over union: A metric and a loss for bounding
  box regression.
\newblock In {\em IEEE Conf. Comput. Vis. Pattern Recog.}, pages 658--666,
  2019.

\bibitem{fixmatchnips2020}
Kihyuk Sohn, David Berthelot, Nicholas Carlini, Zizhao Zhang, Han Zhang,
  Colin~A Raffel, Ekin~Dogus Cubuk, Alexey Kurakin, and Chun-Liang Li.
\newblock Fixmatch: Simplifying semi-supervised learning with consistency and
  confidence.
\newblock {\em Adv. Neural Inform. Process. Syst.}, 33:596--608, 2020.

\bibitem{acrneccv2018}
Chen Sun, Abhinav Shrivastava, Carl Vondrick, Kevin Murphy, Rahul Sukthankar,
  and Cordelia Schmid.
\newblock Actor-centric relation network.
\newblock In {\em ECCV}, pages 318--334, 2018.

\bibitem{sparsercnncvpr2021}
Peize Sun, Rufeng Zhang, Yi Jiang, Tao Kong, Chenfeng Xu, Wei Zhan, Masayoshi
  Tomizuka, Lei Li, Zehuan Yuan, Changhu Wang, et~al.
\newblock Sparse {R-CNN}: End-to-end object detection with learnable proposals.
\newblock In {\em IEEE Conf. Comput. Vis. Pattern Recog.}, pages 14454--14463,
  2021.

\bibitem{meanteachernips2017}
Antti Tarvainen and Harri Valpola.
\newblock Mean teachers are better role models: Weight-averaged consistency
  targets improve semi-supervised deep learning results.
\newblock {\em Adv. Neural Inform. Process. Syst.}, 30, 2017.

\bibitem{fcosiccv2019}
Zhi Tian, Chunhua Shen, Hao Chen, and Tong He.
\newblock {FCOS}: Fully convolutional one-stage object detection.
\newblock In {\em Int. Conf. Comput. Vis.}, pages 9627--9636, 2019.

\bibitem{r2d2aaai2020}
Guo-Hua Wang and Jianxin Wu.
\newblock Repetitive reprediction deep decipher for semi-supervised learning.
\newblock In {\em AAAI}, pages 6170--6177, 2020.

\bibitem{nonlocalcvpr2018}
Xiaolong Wang, Ross Girshick, Abhinav Gupta, and Kaiming He.
\newblock Non-local neural networks.
\newblock In {\em Proceedings of the IEEE conference on computer vision and
  pattern recognition}, pages 7794--7803, 2018.

\bibitem{lfbcvpr2019}
Chao-Yuan Wu, Christoph Feichtenhofer, Haoqi Fan, Kaiming He, Philipp
  Krahenbuhl, and Ross Girshick.
\newblock Long-term feature banks for detailed video understanding.
\newblock In {\em IEEE Conf. Comput. Vis. Pattern Recog.}, pages 284--293,
  2019.

\bibitem{contextrcnneccv2020}
Jianchao Wu, Zhanghui Kuang, Limin Wang, Wayne Zhang, and Gangshan Wu.
\newblock Context-aware {RCNN}: A baseline for action detection in videos.
\newblock In {\em Eur. Conf. Comput. Vis.}, pages 440--456, 2020.

\bibitem{udanips2020}
Qizhe Xie, Zihang Dai, Eduard Hovy, Thang Luong, and Quoc Le.
\newblock Unsupervised data augmentation for consistency training.
\newblock {\em Adv. Neural Inform. Process. Syst.}, 33:6256--6268, 2020.

\bibitem{resnextcvpr2017}
Saining Xie, Ross Girshick, Piotr Doll{\'a}r, Zhuowen Tu, and Kaiming He.
\newblock Aggregated residual transformations for deep neural networks.
\newblock In {\em IEEE Conf. Comput. Vis. Pattern Recog.}, pages 1492--1500,
  2017.

\bibitem{softteachericcv2021}
Mengde Xu, Zheng Zhang, Han Hu, Jianfeng Wang, Lijuan Wang, Fangyun Wei, Xiang
  Bai, and Zicheng Liu.
\newblock End-to-end semi-supervised object detection with soft teacher.
\newblock In {\em Int. Conf. Comput. Vis.}, pages 3060--3069, 2021.

\bibitem{ismtcvpr2021}
Qize Yang, Xihan Wei, Biao Wang, Xian-Sheng Hua, and Lei Zhang.
\newblock Interactive self-training with mean teachers for semi-supervised
  object detection.
\newblock In {\em IEEE Conf. Comput. Vis. Pattern Recog.}, pages 5941--5950,
  2021.

\bibitem{structcvpr2019}
Yubo Zhang, Pavel Tokmakov, Martial Hebert, and Cordelia Schmid.
\newblock A structured model for action detection.
\newblock In {\em IEEE Conf. Comput. Vis. Pattern Recog.}, pages 9975--9984,
  2019.

\bibitem{tubercvpr2022}
Jiaojiao Zhao, Yanyi Zhang, Xinyu Li, Hao Chen, Bing Shuai, Mingze Xu, Chunhui
  Liu, Kaustav Kundu, Yuanjun Xiong, Davide Modolo, et~al.
\newblock {TubeR}: Tubelet transformer for video action detection.
\newblock In {\em IEEE Conf. Comput. Vis. Pattern Recog.}, pages 13598--13607,
  2022.

\bibitem{centernetarxiv2019}
Xingyi Zhou, Dequan Wang, and Philipp Kr{\"a}henb{\"u}hl.
\newblock Objects as points.
\newblock {\em arXiv preprint arXiv:1904.07850}, 2019.

\end{thebibliography}
}

\end{document}